\documentclass[letterpaper]{article} 
\usepackage{aaai2026}
\usepackage{times}  
\usepackage{helvet}  
\usepackage{courier}  
\usepackage[hyphens]{url}  
\usepackage{graphicx} 
\usepackage{natbib}  
\usepackage{caption} 
\usepackage{amsmath}
\usepackage{multirow}
\usepackage{amssymb}
\PassOptionsToPackage{table,xcdraw}{xcolor}
\usepackage{booktabs} 
\usepackage{array}
\usepackage{makecell}
\usepackage{multirow}
\usepackage{algorithm}
\usepackage{algorithmic}

\usepackage{newfloat}
\usepackage{listings}
\DeclareCaptionStyle{ruled}{labelfont=normalfont,labelsep=colon,strut=off} 
\floatstyle{ruled}
\newfloat{listing}{tb}{lst}{}
\floatname{listing}{Listing}
\title{Deformable Attention Graph Representation Learning\\ for Histopathology Whole Slide Image Analysis}

\author {
    Mingxi Fu\textsuperscript{\rm 1}\equalcontrib,
    Xitong Ling\textsuperscript{\rm 1}\equalcontrib,
    Yuxuan Chen\textsuperscript{\rm 1},
    Jiawen Li\textsuperscript{\rm 1},
    Fanglei Fu\textsuperscript{\rm 1},
    Huaitian Yuan\textsuperscript{\rm 1},
    Tian Guan\textsuperscript{1}\textsuperscript{†},
    Yonghong He\textsuperscript{1}\textsuperscript{†},
    Lianghui Zhu\textsuperscript{1}\textsuperscript{}\thanks{Corresponding author.}
}
\affiliations {
    \textsuperscript{\rm 1}Shenzhen International Graduate School, Tsinghua University\\
}

\usepackage{bibentry}

\begin{document}

\maketitle

\begin{abstract}
Accurate classification of Whole Slide Images (WSIs) and Regions of Interest (ROIs) is a fundamental challenge in computational pathology. While mainstream approaches often adopt Multiple Instance Learning (MIL), they struggle to capture the spatial dependencies among tissue structures. Graph Neural Networks (GNNs) have emerged as a solution to model inter-instance relationships, yet most rely on static graph topologies and overlook the physical spatial positions of tissue patches. Moreover, conventional attention mechanisms lack specificity, limiting their ability to focus on structurally relevant regions. In this work, we propose a novel GNN framework with deformable attention for pathology image analysis. We construct a dynamic weighted directed graph based on patch features, where each node aggregates contextual information from its neighbors via attention-weighted edges. Specifically, we incorporate learnable spatial offsets informed by the real coordinates of each patch, enabling the model to adaptively attend to morphologically relevant regions across the slide. This design significantly enhances the contextual field while preserving spatial specificity. Our framework achieves state-of-the-art performance on four benchmark datasets (TCGA-COAD, BRACS, gastric intestinal metaplasia grading, and intestinal ROI classification), demonstrating the power of deformable attention in capturing complex spatial structures in WSIs and ROIs.
\end{abstract}

\begin{figure}[htbp]
\centering
\includegraphics[width=\linewidth]{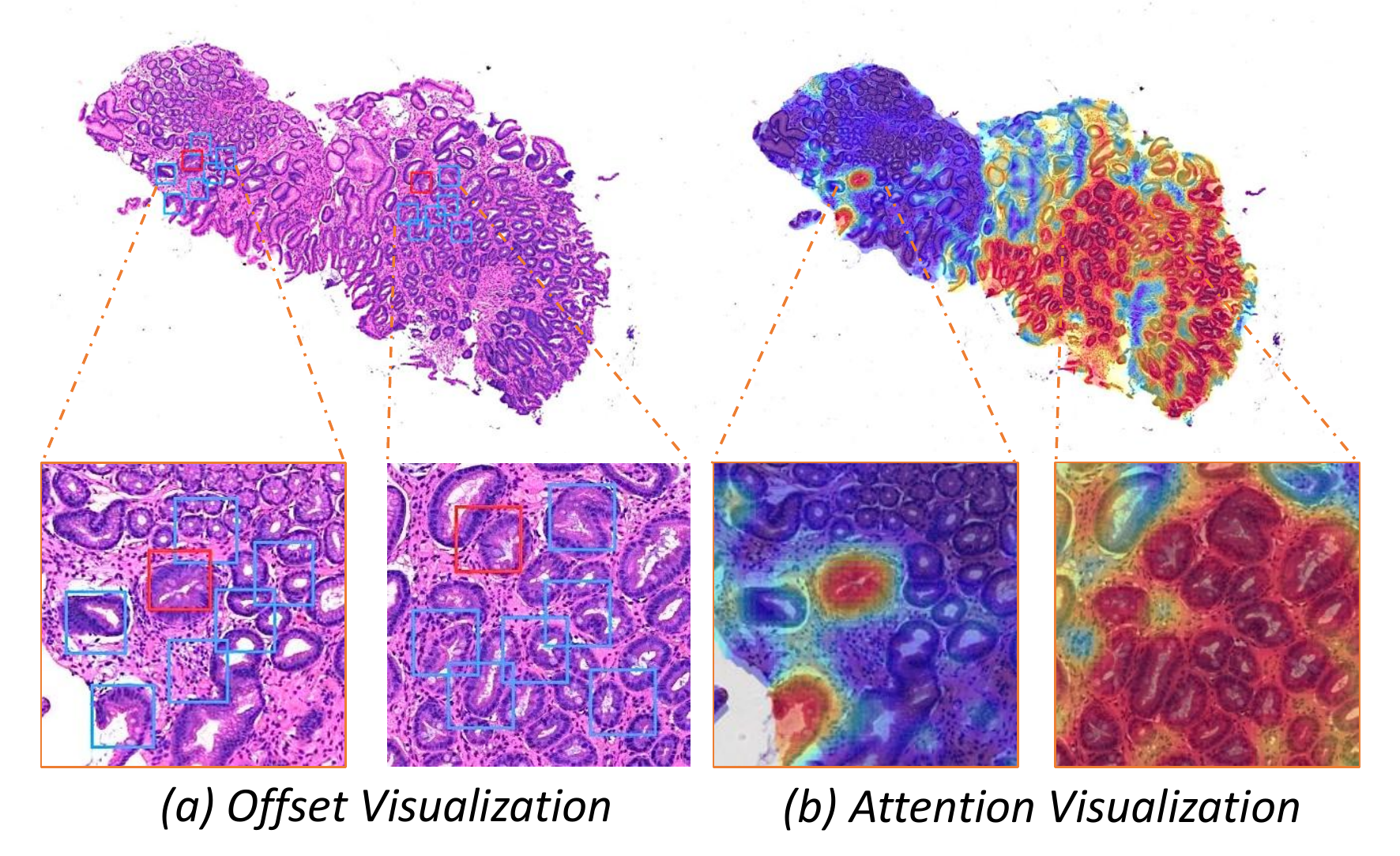}
\caption{Deformable attention graph representation learning with learnable offsets on pathological images.}
\label{fig:offset}
\end{figure}


\section{Introduction}

In computational pathology, the ultra-high resolution of Whole Slide Images (WSIs) makes pixel-level annotation prohibitively time-consuming~\cite{kumar2020whole}, motivating the development of weakly supervised learning techniques~\cite{afonso2024multiple}. Most existing approaches adopt embedding-based Multiple Instance Learning (MIL) frameworks~\cite{li2021dual, yao2020whole, maron1997framework}, which divide a WSI into instance patches and aggregate their features for classification. However, such methods typically neglect the structural correlations between patches, making it difficult to capture the global tissue architecture and spatial dependencies, which are critical for accurate diagnosis~\cite{wang2024advances}.

Graph Neural Networks (GNNs) have recently emerged as a powerful alternative to address this limitation~\cite{ahmedt2022survey}. By modeling the topological relationships between tissue patches, GNNs can capture structural and spatial interactions among key regions~\cite{chan2023histopathology}. Nevertheless, most GNN-based methods rely on static graph structures and disregard the actual spatial coordinates of instances in the WSIs~\cite{li2024dynamic}. This is problematic because spatial context often determines pathological relevance; for example, ductal carcinoma in situ (DCIS) appears within the myoepithelium~\cite{russell2015myoepithelial}, and the loss of surface differentiation in gastric mucosa often indicates malignancy~\cite{khazaaleh2024gastric}.

Transformer-based architectures, empowered by their strong self-attention mechanisms, are capable of modeling interactions among different instances~\cite{shao2021transmil}. However, standard global self-attention suffers from a quadratic computational complexity with respect to the input size, which leads to prohibitive memory and computational costs when processing high-resolution images such as WSIs~\cite{Vaswani2017Attention}. To address this limitation, researchers have proposed various efficient attention mechanisms. For instance, Sparse Attention reduces the complexity through sparse connections~\cite{child2019generating}, Local Window Attention restricts attention computation to local regions to reduce the receptive field~\cite{liu2021swin}, and Pyramid Vision Transformer utilizes a hierarchical pyramid structure to achieve multi-scale representations with reduced overhead~\cite{wang2021pyramid}. Despite their effectiveness, most of these methods rely on fixed or regularized attention patterns, which struggle to adaptively focus on structurally complex or semantically sparse regions in the image. To address this, deformable attention mechanisms~\cite{xia2022vision} introduce learnable sampling offsets that dynamically attend to key spatial positions, achieving a balance between modeling flexibility and computational efficiency. While extensively used in natural image tasks, directly applying deformable attention to gigapixel-scale pathology images remains infeasible due to resolution constraints~\cite{zhu2020deformable}.

In this work, we propose a novel Deformable Attention Graph (DAG) to tackle bag-level classification tasks in Figure~\ref{fig:offset}. We first treat each patch as a graph node and construct a weighted directed graph by learning edge weights based on feature similarity between head and tail nodes. Based on this, we incorporate a deformable attention module guided by the actual spatial coordinates of each patch. This enables the model to dynamically attend to morphologically relevant regions and better adapt to complex tissue structures. We conduct extensive experiments on two public benchmarks (TCGA-COAD for cancer grading and BRACS for breast cancer subtyping) and two in-house clinical datasets (gastric intestinal metaplasia grading and intestinal ROI classification). Compared to state-of-the-art WSI analysis methods and through comprehensive ablation studies, DAG demonstrates superior performance and strong generalization capability in both WSIs and ROIs classification tasks.

\begin{figure}[htbp]
\centering
\includegraphics[width=\linewidth]{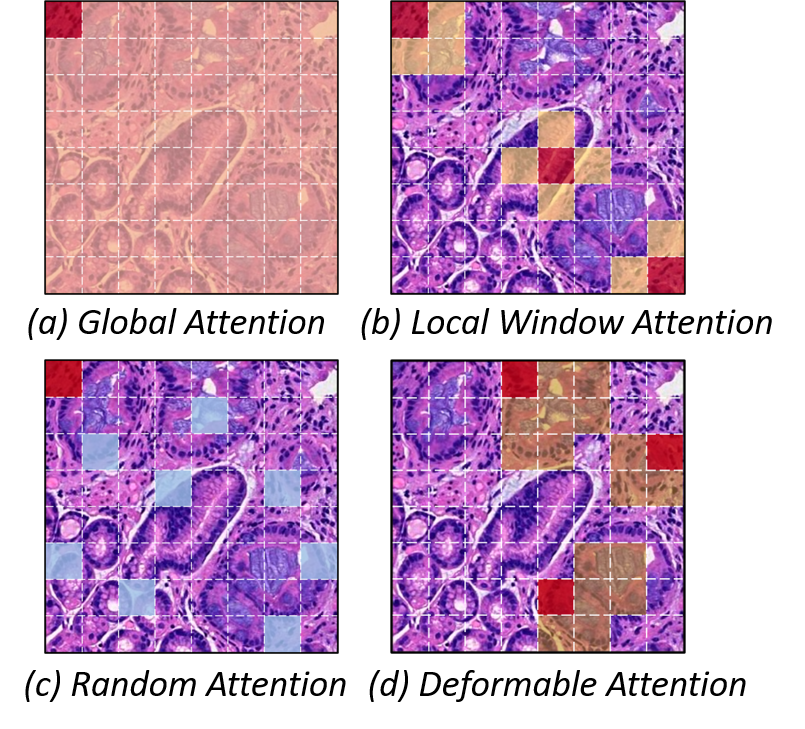}
\caption{Comparison of different attention mechanisms: (a) Global attention spans the entire feature space. (b) Local window attention focuses on the current position and its neighboring regions. (c) Random attention establishes connections between randomly positions. (d) Deformable attention focuses on morphological contours dynamically.}
\label{fig:attention1}
\end{figure}

\begin{figure*}[htbp]
\centering
\includegraphics[width=\textwidth]{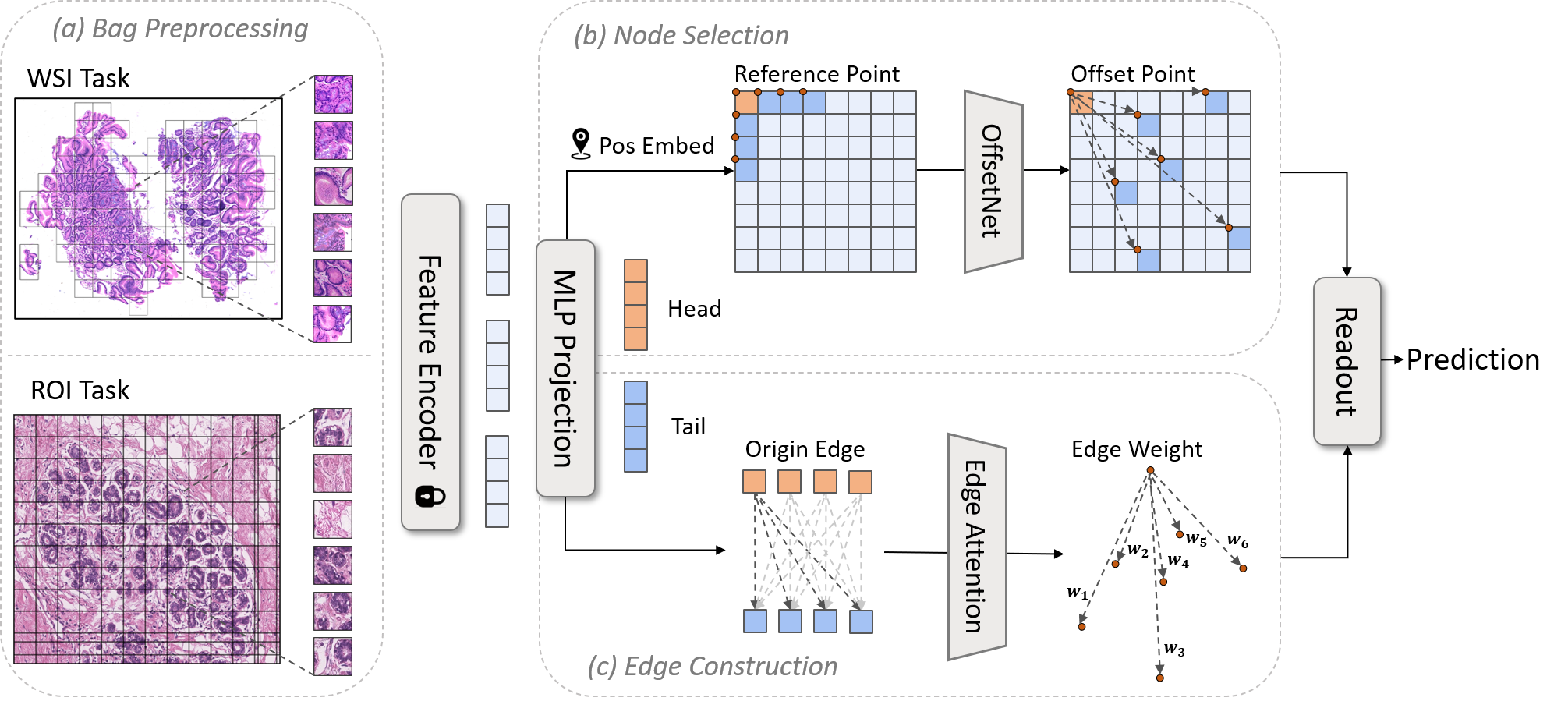}
\caption{Overall process: (a) The preprocessing of WSIs and ROIs. (b) Node selection through offsetnet in graph network. (c) Dynamic weighted edge construction in graph network.}
\label{fig:model}
\end{figure*}

\section{Related Work}
\subsection{Attention Mechanisms in WSIs Analysis}
MIL techniques with different attention mechanisms in Figure~\ref{fig:attention1} have achieved remarkable success in weakly supervised WSI analysis by adaptively learning instance weights and prioritizing diagnostically informative regions~\cite{gadermayr2024multiple}. ABMIL uses the global attention to each patch for key instance aggregation, but it ignores the spatial positions and structural relationships among patches, limiting its ability to capture local-global tissue dependencies~\cite{ilse2018attention}. Li et al.~\cite{li2022simvit} introduced window-based attention, reducing the computational complexity of global attention and enhancing efficiency. Furthermore, the BigBird model~\cite{zaheer2020big} combines sliding window attention with sparse global and random attention patterns, enabling efficient modeling of long-range dependencies. 

To overcome these limitations, deformable attention mechanisms have been proposed~\cite{zhu2019deformable}. By introducing learnable offsets, these mechanisms enable the model to dynamically attend to key locations and better model structural contours~\cite{li2021dt}. Recently, such mechanisms have demonstrated strong performance in vision transformer architectures. For example, Xia et al. proposed the Deformable Attention Transformer (DAT), which achieved outstanding results on image classification and dense prediction tasks~\cite{xia2022vision}. Moreover, an enhanced version named DAT++ is built upon the DAT, achieving state-of-the-art performance on various classification and segmentation tasks in natural image domains~\cite{xia2023dat++}. However, existing deformable attention methods are primarily designed for natural images and typically operate by locating offset sampling points directly on the input image. In the case of ultra-high-resolution whole slide images (WSIs), it is infeasible to put the entire WSIs into the network for offset computation due to memory and computational constraints.

\subsection{Graph Representation in WSIs Analysis}

In recent years, GNNs have demonstrated great potential in histopathological image classification tasks~\cite{jaume2021quantifying}. Unlike traditional MIL methods, GNNs construct graph structures to effectively model the spatial and structural relationships between cells or tissue regions. Early approaches typically relied on static graph structures. For example, ABMIL-GCN embeds patch-level features into a graph and leverages attention mechanisms to capture context-aware information~\cite{liang2023interpretable}. However, it often neglects explicit structural dependencies between patch nodes. To better model spatial hierarchies, recent methods have introduced hierarchical representation strategies. Wang et al. proposed a Connectivity-Aware Graph Transformer (CGT), which enhances graph representation by integrating connection information into the node features of each graph transformer layer, significantly improving breast cancer subtype classification performance~\cite{wang2024breast}. HACT-Net introduced a hierarchical GNN that models cell-to-tissue level structures, leading to improved multi-class classification of breast cancer subtypes~\cite{pati2020hact}. In terms of modeling long-range dependencies, Su et al. combined GNNs and transformers in HAT-Net to better capture both structural and contextual information, achieving strong performance on colorectal cancer grading tasks~\cite{su2021hat}.

To overcome the limitations of static graph structures in modeling distant structural dependencies, WiKG reformulates WSIs as knowledge graphs and constructs dynamic graph representations to enable end-to-end learning~\cite{li2024dynamic}. Similarly, Chen et al. proposed a dynamic hypergraph neural network (DyHG) using connect multiple nodes to overcome the constraints of fixed graph topologies~\cite{chen2025dynamic}. Despite these advances, most existing methods overlook the actual physical locations of patches during graph construction and lack mechanisms to model edge weights. These limitations hinder the expressive power of GNNs when dealing with the complex spatial structures inherent in WSIs.

\section{Methodology}
\subsection{Deformable Graph Construction}

Given a WSI, we first segment the foreground regions using the OTSU thresholding method in Figure~\ref{fig:model}. Then, we apply a sliding window operation to divide the WSIs into non-overlapping patches $P = \{p_1, p_2, \dots, p_N\}$, where each patch is regarded as a graph node. We extract features from each patch using pretrained pathology models (e.g., UNI~\cite{chen2024towards}). Each patch also has its real-world coordinate $coord = \{(x_1, y_1), (x_2, y_2), \dots, (x_N, y_N)\}$, where $c_i = (x_i, y_i)$ denotes the location of the $i$-th patch. Based on this, we construct a directed graph where each node has both a head and a tail representation. The head node focuses on discovering the relationships between itself and other patches, while the tail node evaluates its own contribution to the head nodes.
\begin{align}
    h_i = W_h f(P), \quad t_i = W_t f(P),
\end{align}
where $h_i$ and $t_i$ respectively denote the head and tail em
beddings of patch $i$. Traditional deformable attention typically samples from a uniformly distributed reference grid and learns offsets for each point, which are used to sample neighboring positions. Inspired by this idea, we transform the entire WSI into a set of reference points, then input each head node into a lightweight network $\mathcal{O}_{\text{offset}}(\cdot)$ to generate $K$ pixel-level offsets $O_i = \{o_{i,1}, o_{i,2}, \dots, o_{i,K}\}$:
\begin{align}
    O_i = \{i \in \mathcal{V}: o_i = \mathcal{O}_{\text{offset}}(h_i)\},
\end{align}
where $\mathcal{V}$ is the set of patch nodes and $|\mathcal{V}| = N$, $|O_i| = K$. To dynamically normalize these offsets for better neighbor sampling, we apply the following transformation:
\begin{align}
    O_i = O_i \times S \times \sqrt{N} \times \sigma(\alpha),
\end{align}
where $S$ denotes the spatial stride used for feature extraction, which maps the offset from the feature space back to the original image space. Let $N$ be the number of patches in each WSI. To account for different WSI sizes, we normalize the offset using the patch number $N$ and use a learnable parameter $\alpha$ activated through a sigmoid function to adjust the offset scale dynamically.

Previous studies often construct implicit structures purely based on relative patch locations while ignoring the absolute spatial positions of patches. To address this, we incorporate the actual spatial coordinates of each patch to generate deformable query positions as follows:
\begin{equation}
q_{i,k} = c_i + o_{i,k}, \quad k = 1, \dots, K,
\end{equation}
where $q_{i,k}$ represents the position of the $k$-th query point of the $i$-th patch after applying the offset, which defines the dynamically updated attention field. Next, we calculate the Euclidean distance between each query point $q_{i,k}$ and all real patch coordinates $c_j$, and find the closest real patch $c_j$ as its neighbor:
\begin{equation}
D_{i,k} = \arg \min_{j \in \{1, \dots, N\}} \| q_{i,k} - c_j \|_2^2,
\end{equation}
where $i$ denotes the head node, $k$ is the query index, and $j$ is the tail node. For each of the $K$ queries of node $i$, we obtain a neighbor set as:
\begin{equation}
n_{i,k} = t_{D_{i,k}}.
\end{equation}

Finally, the full set of dynamically sampled neighbors for node $i$ under the deformable graph structure is $N_i = \{n_{i,1}, n_{i,2}, \dots, n_{i,K} \}$.

\subsection{Dynamic Edge Weights Learning}
To fully leverage the offset information gained from the graph, we design an attention mechanism that weights neighbors using both the head and tail node features as well as the learned offsets. Specifically, we first expand the head node feature $h_i$ via linear transformation. Then, for each pair of head node and neighbor node, we compute their cosine similarity as:
\begin{equation}
s_{i,k} = \frac{(h_i \cdot n_{i,k})^\top}{\|h_i\|_2 \|n_{i,k}\|_2}.
\end{equation}

We then apply softmax normalization over the $K$ neighbors of node $i$:
\begin{equation}
\alpha_{i,k} = \frac{\exp(s_{i,k})}{\sum_{k=1}^{K} \exp(s_{i,k})},
\end{equation}
where $\alpha_{i,k}$ denotes the attention weight for the $k$-th edge of node $i$, indicating the influence strength of the $k$-th neighbor node. It guides the message passing from the tail node to the head node. Besides, we use a gating mechanism to fuse head and neighbor node features, while also learning the difference between them:
\begin{equation}
u_{i,k} = \tanh(h_i + \alpha_{i,k} \cdot n_{i,k}),
\end{equation}
\begin{equation}
e_i = \text{Softmax}(u_{i,k} \cdot n_{i,k}),
\end{equation}
where $e_i$ represents the aggregated neighbors information after  updated attention. We then fuse $e_i$ with the original head feature to form an updated representation for the head node, which will be used for downstream classification. To further enhance information flow between nodes, we adopt a dual-channel residual fusion mechanism:
\begin{equation}
h_i = \sigma_1(W_1(h_i + e_i)) + \sigma_2(W_2(h_i \odot e_i)),
\end{equation}
where $\sigma$ represents an activation function such as LeakyReLU, and $W_1$, $W_2$ are learnable projection matrices. Finally, a global readout function is used to aggregate patch-level representations into slide-level representation for classification:
\begin{equation}
\hat{Y} = \text{Softmax}(\text{Readout}(G)).
\end{equation}

The $\text{Readout}$ function can be instantiated as global average pooling, max pooling, or attention-based pooling, where $\hat{Y}$ is the predicted class probability for the WSI.

\begin{table*}[htbp]
\centering
\small

\renewcommand{\arraystretch}{1.4}
\setlength{\tabcolsep}{4pt}
\begin{tabular}{
    >{\centering\arraybackslash}p{1.8cm}
    *{12}{>{\centering\arraybackslash}p{1.0cm}}
}
\toprule
\multirow{2}{*}{\textbf{Method}} &
\multicolumn{3}{c}{\textbf{TCGA-COAD}} &
\multicolumn{3}{c}{\textbf{Gastritis-IM}} &
\multicolumn{3}{c}{\textbf{BRACS}} &
\multicolumn{3}{c}{\textbf{Intestine}} \\
\cmidrule(lr){2-4} \cmidrule(lr){5-7} \cmidrule(lr){8-10} \cmidrule(lr){11-13}
& ACC & AUC & F1 & ACC & AUC & F1 & ACC & AUC & F1 & ACC & AUC & F1 \\
\midrule
ABMIL & 86.24\textsubscript{1.40} & \underline{95.38\textsubscript{0.24}} & 84.39\textsubscript{1.11} & 78.36\textsubscript{3.07} & 94.12\textsubscript{0.86} & 75.71\textsubscript{3.27} & 57.60\textsubscript{1.19} & 90.05\textsubscript{0.30} & 56.56\textsubscript{1.47} & 94.67\textsubscript{0.40} & 99.36\textsubscript{0.12} & 94.62\textsubscript{0.41} \\
AMDMIL & 85.81\textsubscript{2.07} & 95.38\textsubscript{0.15} & 84.21\textsubscript{1.59} & 
79.11\textsubscript{3.28} & 94.55\textsubscript{1.73} & \underline{79.10\textsubscript{2.53}} &
58.47\textsubscript{2.36} & 89.14\textsubscript{2.39} & 57.03\textsubscript{2.28} &
96.16\textsubscript{0.48} & \underline{99.36\textsubscript{0.20}} & 96.15\textsubscript{0.48} \\
CLAM-SB & 86.67\textsubscript{0.96} & 93.65\textsubscript{0.34} & 84.00\textsubscript{1.06} &
79.47\textsubscript{1.42} & \underline{94.98\textsubscript{0.54}} & 78.58\textsubscript{1.20} &
56.22\textsubscript{2.62} & 89.24\textsubscript{0.68} & 55.86\textsubscript{2.02} &
96.14\textsubscript{1.24} & 99.20\textsubscript{0.56} & 96.14\textsubscript{0.53} \\
FRMIL & 84.09\textsubscript{2.07} & 92.38\textsubscript{1.11} & 80.26\textsubscript{2.23} &
73.98\textsubscript{1.77} & 90.50\textsubscript{1.75} & 72.96\textsubscript{2.13} &
55.96\textsubscript{1.90} & 84.01\textsubscript{3.39} & 55.72\textsubscript{1.77} &
83.58\textsubscript{1.64} & 94.15\textsubscript{0.86} & 83.44\textsubscript{1.63} \\
TransMIL & 86.45\textsubscript{1.95} & 95.07\textsubscript{0.98} & 83.70\textsubscript{2.46} &
77.84\textsubscript{3.11} & 94.50\textsubscript{0.69} & 76.77\textsubscript{2.08} &
55.38\textsubscript{1.50} & 89.35\textsubscript{0.70} & 55.21\textsubscript{1.34} &
88.28\textsubscript{1.77} & 98.12\textsubscript{0.32} & 87.90\textsubscript{1.97} \\
DM-GNN & 86.67\textsubscript{1.23} & 94.40\textsubscript{0.66} & 84.36\textsubscript{1.20} &
78.15\textsubscript{1.05} & 93.99\textsubscript{0.88} & 77.48\textsubscript{1.82} &
57.27\textsubscript{1.34} & 89.26\textsubscript{1.65} & 55.68\textsubscript{1.34} &
\underline{96.22\textsubscript{0.48}} & 99.30\textsubscript{0.11} & \underline{96.26\textsubscript{0.49}} \\
DyHG & \underline{86.67\textsubscript{1.63}} & 95.18\textsubscript{0.67} & \underline{84.42\textsubscript{2.39}} &
\underline{79.59\textsubscript{2.67}} & 93.42\textsubscript{2.81} & 78.95\textsubscript{1.44} &
\underline{58.91\textsubscript{1.47}} & \underline{90.13\textsubscript{0.22}} & \underline{57.79\textsubscript{1.50}} &
96.21\textsubscript{0.63} & 99.35\textsubscript{0.32} & 96.19\textsubscript{0.63} \\
Patch-GCN & 85.81\textsubscript{0.90} & 94.52\textsubscript{1.30} & 81.54\textsubscript{3.23} &
73.98\textsubscript{2.27} & 91.60\textsubscript{1.56} & 71.12\textsubscript{2.65} &
54.47\textsubscript{1.96} & 86.07\textsubscript{0.91} & 54.19\textsubscript{1.52} &
89.92\textsubscript{1.39} & 97.95\textsubscript{0.20} & 89.79\textsubscript{1.43} \\
WiKG & 85.81\textsubscript{3.26} & 94.19\textsubscript{1.20} & 83.72\textsubscript{2.83} &
76.62\textsubscript{5.65} & 93.67\textsubscript{2.02} & 74.95\textsubscript{4.97} &
57.78\textsubscript{2.85} & 89.95\textsubscript{0.67} & 56.89\textsubscript{2.93} &
94.69\textsubscript{0.90} & 99.21\textsubscript{0.20} & 94.64\textsubscript{0.91} \\
DAG & \textbf{87.31\textsubscript{1.59}} & \textbf{95.73\textsubscript{0.56}} & \textbf{84.89\textsubscript{1.50}} &
\textbf{80.39\textsubscript{2.60}} & \textbf{95.24\textsubscript{0.48}} & \textbf{79.27\textsubscript{3.48}} &
\textbf{59.67\textsubscript{1.34}} & \textbf{90.20\textsubscript{0.81}} & \textbf{58.11\textsubscript{1.17}} &
\textbf{96.36\textsubscript{1.42}} & \textbf{99.46\textsubscript{0.20}} & \textbf{96.33\textsubscript{1.46}} \\
\bottomrule
\end{tabular}
\caption{Performance of DAG on TCGA-COAD, Gastric intestinal metaplasia, BRACS, and Intestinal cancer datasets.}
\label{tab:sota}

\end{table*}

\section{EXPERIMENT}
\subsection{Datasets}

We evaluate our proposed method on four datasets, covering both WSIs and ROIs level classification tasks. Specifically, we use two WSI-level datasets: the publicly available TCGA-COAD and a private gastric intestinal metaplasia grading dataset; and two ROI-level datasets: the publicly available BRACS breast cancer subtyping dataset and a private intestinal cancer classification dataset.\\
\textbf{TCGA-COAD (Public WSI dataset).} This dataset includes 465 WSIs from the TCGA Colon Adenocarcinoma (TCGA-COAD) cohort. The cases are categorized into four classes: Adenomas and Adenocarcinomas (388), Cystic, Mucinous and Serous Neoplasms (60), Complex Epithelial Neoplasms (11), and Epithelial Neoplasms, NOS (6).\\
\textbf{Gastric intestinal metaplasia (Private WSI dataset).} This dataset is collected from the Second Affiliated Hospital of Southern University of Science and Technology. It contains 984 gastric WSIs diagnosed with different grades of intestinal metaplasia, including 309 slides without metaplasia, 299 with mild metaplasia, 107 with moderate metaplasia, and 269 with severe metaplasia. \\
\textbf{BRACS (Public ROI dataset).} This dataset contains 4492 ROIs related to breast cancer subtypes. It covers seven categories: benign (833), usual ductal hyperplasia (UDH, 506), flat epithelial atypia (FEA, 754), atypical ductal hyperplasia (ADH, 503), ductal carcinoma in situ (DCIS, 783), invasive carcinoma (647), and normal tissue (466). \\
\textbf{Intestinal cancer classification (Private ROI dataset).} This dataset consists of ROIs extracted from WSIs of intestinal tissues provided by the Chongqing University Affiliated Three Gorges Hospital. It includes a total of 9381 ROIs, categorized into four classes: cancer (3591), high-grade intraepithelial neoplasia (767), low-grade intraepithelial neoplasia (2982), and non-tumor (2041).\\
We report the mean and standard deviation of three standard evaluation metrics: accuracy (ACC), weighted F1-score, and area under the ROC curve (AUCROC). 

\subsection{Implementation Details}

During the pre-processing stage, WSI-level datasets are partitioned into non-overlapping patches of size $1024\times1024$ at $10\times$ magnification, while ROI-level datasets are divided into non-overlapping patches of $256\times256$ pixels. All experiments are conducted using consistent hyperparameters on a workstation equipped with NVIDIA RTX A100 GPUs. We employ the UNI model as the feature encoder, which is pre-trained on histopathological images~\cite{chen2024towards}. All tasks are evaluated using 5-fold cross-validation, with the training, validation, and test sets split in a ratio of 7:2:1. During training, we use the cross-entropy loss, the Adam optimizer with a base learning rate of 0.001, and a weight decay coefficient of 1e-5. Training is conducted for 70 epochs, and model performance is monitored using validation accuracy. Early stopping is applied with a patience of 30 epochs.

\subsection{Comparison with State-of-the-Art Methods}

In this study, we present the experimental results of our proposed DAG framework on four datasets and compare it against both traditional MIL methods and graph-based approaches. Specifically, we compare with (1) ABMIL~\cite{ilse2018attention}, a classical MIL method that aggregates instance features via attention to generate bag-level embeddings; (2) AMDMIL~\cite{ling2024agent}, which introduces an agent aggregator with a mask denoising mechanism for WSI analysis; (3) CLAM-SB~\cite{lu2021data}, a gated attention-based MIL framework optimized using a clustering-constrained loss to improve instance selection; (4) FRMIL~\cite{chikontwe2024fr}, which recalibrates WSI bag distributions by leveraging statistics from critical instances; (5) TransMIL~\cite{shao2021transmil}, a transformer-based MIL method incorporating multiscale position encodings to capture inter-instance dependencies; (6) DM-GNN~\cite{wang2024dual}, a dual-stream graph network equipped with affinity-guided attention recalibration for robust global graph representation; (7) DyHG~\cite{chen2025dynamic}, a dynamic hypergraph network that constructs hyperedges via Gumbel-Softmax-based nonlinear transformations; (8) Patch-GCN~\cite{chen2021whole}, a hierarchical graph model designed for WSIs analysis using global attention pooling; and (9) WIKG~\cite{li2024dynamic}, which conceptualizes WSIs as dynamic knowledge graphs for end-to-end graph learning. Our DAG framework consistently outperforms these baselines across WSIs and ROIs level classification tasks, demonstrating its effectiveness in modeling weighted spatial relationships in histopathological images.

As shown in Table~\ref{tab:sota}, our proposed DAG model achieves the best performance across all four tasks, including two WSI-level tasks (TCGA-COAD and Gastritis-IM) and two ROI-level tasks (Intestine and BRACS). Compared to the second-best methods, DAG has improvements in accuracy, with gains of 0.64\% on TCGA-COAD, 0.8\% on Gastritis-IM, 0.76\% on BRACS, and 0.14\% on Intestine. These results demonstrate that DAG can effectively adapt to the complex spatial distribution of WSIs while capturing fine-grained structural variations.

\begin{figure*}[htbp]
\centering
\includegraphics[width=\textwidth]{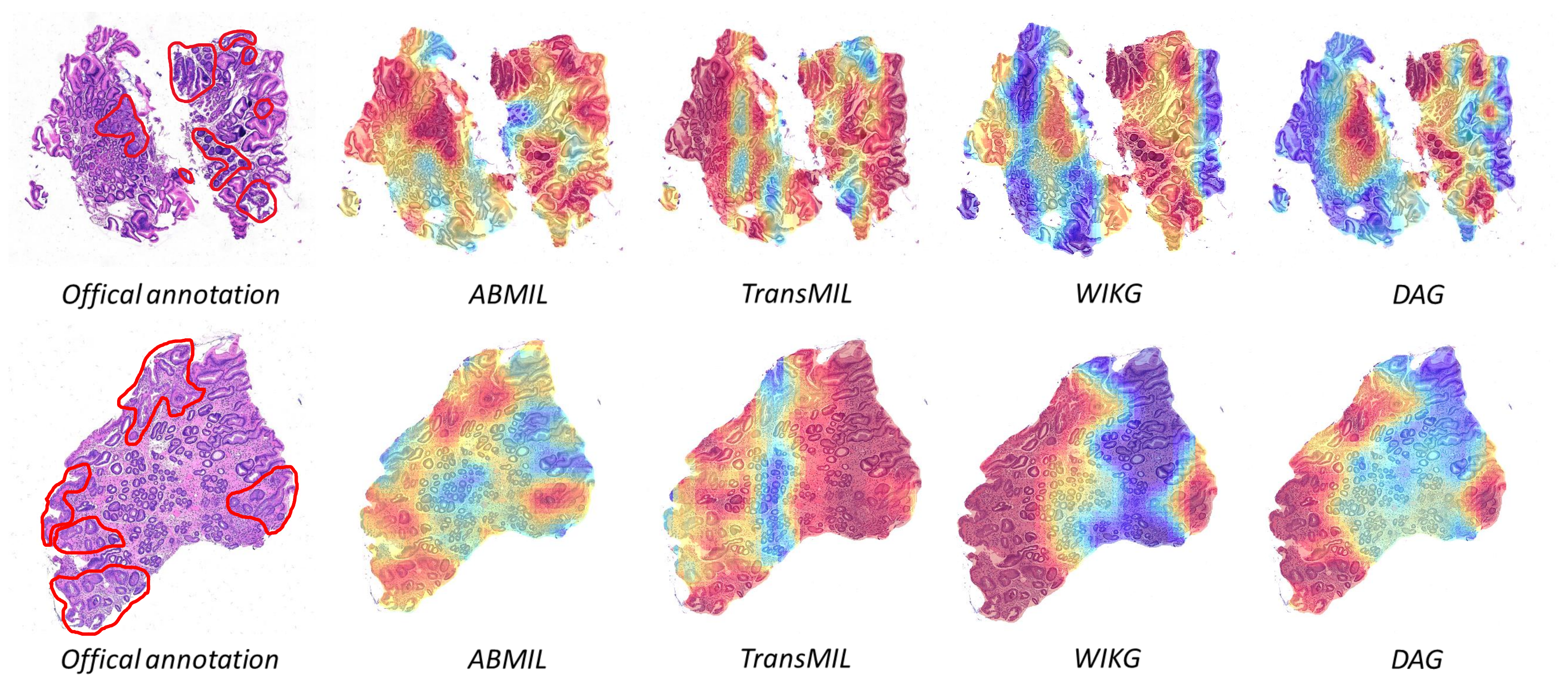}
\caption{Visualization of DAG Attention Distribution Compared to Official Annotations on Gastric intestinal metaplasia dataset.}
\label{fig:visual}
\end{figure*}

\subsection{Interpretability Analysis}

To further evaluate the advantages of our proposed DAG method in lesion localization, we conducted a visual comparison on representative WSIs of Gastric intestinal metaplasia dataset with several mainstream MIL and graph-based models, including ABMIL~\cite{ilse2018attention}, TransMIL~\cite{shao2021transmil}, WIKG~\cite{li2024dynamic}, and DAG, as shown in Figure~\ref{fig:visual}. The red annotations in the 'Official annotation' column indicate the ground-truth lesion regions provided by pathologists.

As illustrated in Figure~\ref{fig:visual}, DAG generates the most accurate attention heatmaps for intestinal metaplasia areas. All three comparative models exhibited a tendency to either concentrate on substantial non-intestinal metaplasia regions or neglect specific intestinal metaplasia foci. In contrast, DAG demonstrates a good concordance with pathological annotations, while maintaining robust capability in concentrating on both adjacent lesion clusters and distally scattered multifocal lesions. These findings suggest DAG achieves a strong balance between coarse-grained lesion localization and fine-grained boundary recognition.

\subsection{Ablation Study}

\textbf{Effectiveness of key components of DAG.} We conduct comprehensive ablation studies to investigate the effectiveness of the key components in our proposed method, namely the offset prediction module (Offset), the edge weight module (Weight), and the spatial coordinate module (Coords). The results evaluated on four datasets are summarized in Table~\ref{tab:ablation}.

\textbf{(1) Offset Module.} The learnable offset module empowers the model to dynamically adjust the receptive field for neighboring patch selection, enabling enhanced capture of structural layouts and improved identification of lesion regions across WSIs. As evidenced by ablation studies on four datasets, incorporating the offset module consistently yields superior performance compared to its absence. These results demonstrate the module’s effectiveness in modeling fine-grained tissue structures and strengthening the model's capacity to focus on critical pathological regions.

\textbf{(2) Edge Weight Module.} The edge weight module dynamically computes and updates attention weights between nodes and offset, indicating the relative importance of each neighbor during aggregation. In the ablation results across four datasets, removing the edge weights clearly leads to worse performance. This suggests that edge attention weights are essential for emphasizing highly correlated neighbors, which is necessary for accurate recognition of complex pathological structures.

\textbf{(3) Coordinate Module.} The coordinate module provides the model with absolute spatial positions of individual patches, facilitating the modeling of spatial dependencies and avoiding selecting distant unrelated patches as neighbors. Across all ablation studies, the integration of coordinate information consistently improves performance, highlighting its importance in holding tissue structural relationships and locating lesion areas.

\begin{table}[htbp]
\centering
\footnotesize  
\renewcommand{\arraystretch}{1.1}

\setlength{\tabcolsep}{3pt}
\begin{tabular}{c c c c c c c}
\toprule
\multirow{2.5}{*}{\textbf{Dataset}} & 
\multicolumn{3}{c}{\textbf{Component}} &  
\multirow{2.5}{*}{\textbf{ACC}} &
\multirow{2.5}{*}{\textbf{AUC}} &
\multirow{2.5}{*}{\textbf{F1}} \\
\cmidrule(lr){2-4}
& Offset & Weight & Coords \\
\midrule
\multirow{4}{*}{COAD}
& \checkmark & \checkmark &        & 84.73\textsubscript{3.08} & 94.39\textsubscript{0.55} & 82.47\textsubscript{2.73} \\
& \checkmark &            & \checkmark & 85.16\textsubscript{2.45} & \underline{94.92\textsubscript{1.04}} & 81.35\textsubscript{3.80} \\
&           & \checkmark & \checkmark & \underline{85.38\textsubscript{2.36}} & 94.82\textsubscript{0.24} & \underline{82.74\textsubscript{2.38}} \\
& \checkmark & \checkmark & \checkmark & \textbf{87.31\textsubscript{1.59}} & \textbf{95.73\textsubscript{0.56}} & \textbf{84.89\textsubscript{1.50}} \\
\midrule
\multirow{4}{*}{IM}
& \checkmark & \checkmark &        & 77.85\textsubscript{2.53} & 92.80\textsubscript{2.73} & 76.75\textsubscript{3.21} \\
& \checkmark &            & \checkmark & \underline{78.76\textsubscript{4.52}} & \underline{93.88\textsubscript{1.92}} & \underline{77.60\textsubscript{4.29}} \\
&           & \checkmark & \checkmark & 78.04\textsubscript{3.56} & 93.82\textsubscript{1.58} & 77.30\textsubscript{1.66} \\
& \checkmark & \checkmark & \checkmark &
\textbf{80.39\textsubscript{2.60}} & \textbf{95.24\textsubscript{0.48}} & \textbf{79.27\textsubscript{3.44}} \\
\midrule
\multirow{4}{*}{BRACS}
& \checkmark & \checkmark &        & 57.35\textsubscript{1.50} & 89.21\textsubscript{0.69} & 56.94\textsubscript{1.46} \\
& \checkmark &            & \checkmark & \underline{58.76\textsubscript{1.90}} & \underline{89.34\textsubscript{1.15}} & \underline{57.64\textsubscript{1.91}} \\
&           & \checkmark & \checkmark & 56.15\textsubscript{1.46} & 88.67\textsubscript{1.54} & 54.43\textsubscript{2.90} \\
& \checkmark & \checkmark & \checkmark &
\textbf{59.67\textsubscript{1.34}} & \textbf{90.20\textsubscript{0.81}} & \textbf{58.11\textsubscript{1.17}} \\
\midrule
\multirow{4}{*}{Intestine}
& \checkmark & \checkmark &        & 95.39\textsubscript{0.10} & \underline{99.45\textsubscript{0.07}} & 95.39\textsubscript{1.00} \\
& \checkmark &            & \checkmark & 93.92\textsubscript{1.23} & 99.09\textsubscript{0.14} & 93.86\textsubscript{1.32} \\
&           & \checkmark & \checkmark & \underline{95.48\textsubscript{1.30}} & 99.42\textsubscript{0.23} & \underline{95.47\textsubscript{1.26}} \\
& \checkmark & \checkmark & \checkmark &
\textbf{96.36\textsubscript{1.42}} & \textbf{99.46\textsubscript{0.20}} & \textbf{96.33\textsubscript{1.46}} \\
\bottomrule
\end{tabular}
\caption{Ablation study on the effectiveness of DAG components.}
\label{tab:ablation}
\end{table}

\begin{figure}[htbp]
\centering
\includegraphics[width=\linewidth]{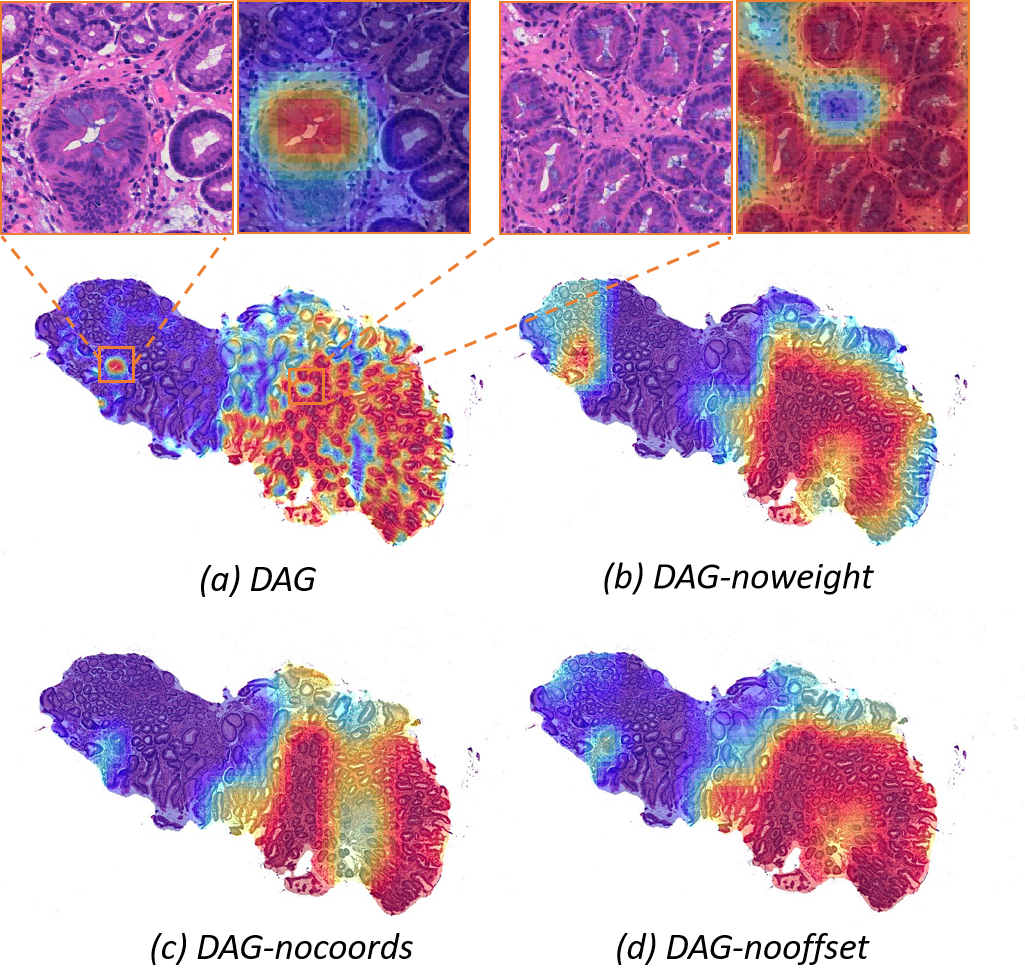}
\caption{Attention distribution of different components: (a) Deformable attention graph with weight, coords and offset three modules. (b) Deformable attention graph without weight module. (c) Deformable attention graph without coords module. (d) Deformable attention graph without offset module.}
\label{fig:attention}
\end{figure}

Figure~\ref{fig:attention} presents the visualization of the ablation results. For the intestinal metaplasia (IM) grading task in gastritis, pathologists often focus on gland regions during diagnosis. The visualization from our proposed DAG demonstrates its ability to precisely distinguish each intestinal metaplasia gland and accurately delineate the gland boundaries. Moreover, thanks to the dynamically learnable offset mechanism, DAG is capable of performing long-range attention, effectively capturing distant intestinal metaplasia regions and localizing them with high accuracy. In addition, Figure~\ref{fig:attention} also illustrates the critical roles of the three key modules in DAG. Without the edge weight module, the model fails to assess the importance of neighbor nodes during graph construction, resulting in incorrect attention allocation. Without spatial coordinates, the model loses awareness of distant lesion regions and fails to capture the underlying structural dependencies among patches. Besides, removing the offset module results in coarse-grained attention patterns. The model can roughly identify lesion regions but fail to recognize their precise morphological boundaries.

\textbf{Effectiveness of Hyperparameter.} We conduct our ablation experiments on Hyperparameter \textit{topk} and \textit{stride}. The hyperparameter \textit{topk} determines the number of neighboring nodes selected during the dynamic graph construction process. The hyperparameter \textit{stride} controls the spatial sampling range during the dynamic graph construction.

\begin{figure}[htbp]
\centering
\includegraphics[width=\linewidth]{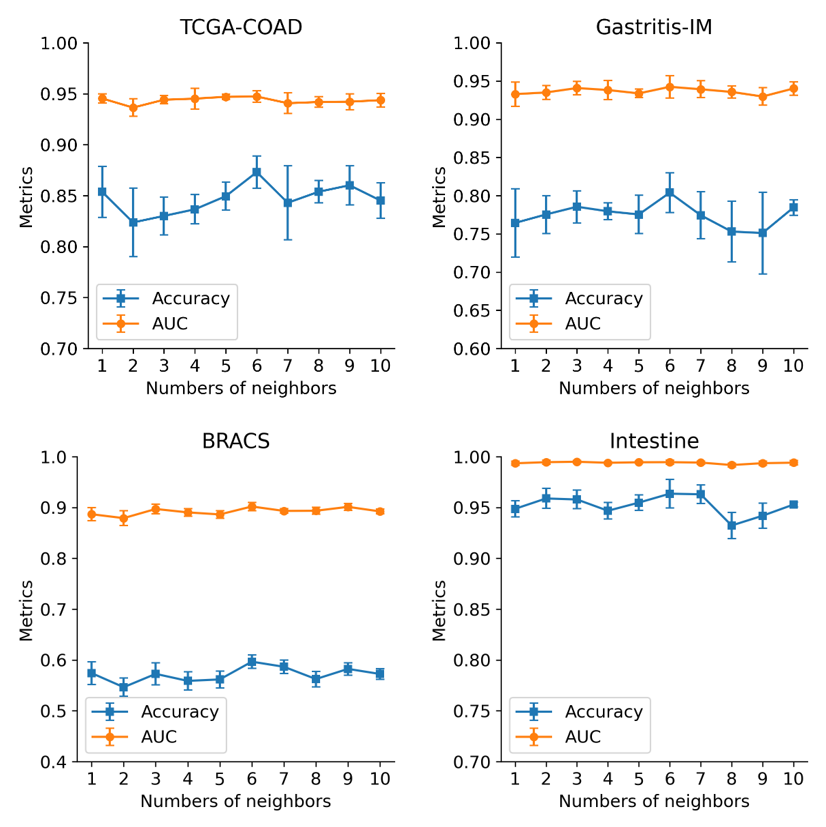}
\caption{Classification results of AUC and ACC scores with different numbers of neighbor nodes on four datasets.}
\label{fig:topk}
\end{figure}

\textbf{(1) Influence of Hyperparameter \textit{topk}.} Figure~\ref{fig:topk} shows that a moderate number of neighbors achieves a good balance between computational efficiency and representational power. If \textit{topk} is too small, the model captures limited information, potentially missing critical pathological structures. Conversely, setting \textit{topk} too large may aggregate excessive irrelevant or noisy information, degrading the discriminative capability of the model.

\textbf{(2) Influence of Hyperparameter \textit{stride}.} Table~\ref{tab:stride} presents the classification accuracy under different \textit{stride} settings across four datasets. The results show that each dataset achieves optimal performance at different stride values. A smaller \textit{stride} allows the model to focus on a more localized spatial region, thereby capturing fine-grained pathological structures more effectively. For example, in the intestinal metaplasia classification task, glandular structures are critical factors for diagnosing, and the model performs better with smaller strides. In contrast, a larger \textit{stride} enables the model to perceive broader spatial dependencies and capture more global contextual information. For instance, the BRACS classification task emphasizes overall structural patterns of the tissue, thus achieving superior performance with larger strides.

\begin{table}[htbp]
\centering
\footnotesize
\renewcommand{\arraystretch}{1.2}

\setlength{\tabcolsep}{6pt}
\begin{tabular}{ccccc}
\toprule
\textbf{Stride} & \textbf{BRACS} & \textbf{COAD} & \textbf{IM} & \textbf{Intestine} \\
\midrule
64   & \underline{57.27\textsubscript{1.28}} & \underline{85.81\textsubscript{1.77}} & 78.45\textsubscript{2.01} & \textbf{95.30\textsubscript{0.93}} \\
128  & 56.47\textsubscript{2.23} & 85.59\textsubscript{1.63} & \textbf{80.39\textsubscript{2.60}} & 94.16\textsubscript{2.23} \\
256  & 56.04\textsubscript{2.45} & 84.95\textsubscript{1.32} & 78.66\textsubscript{2.91} & 94.13\textsubscript{0.62} \\
512  & 56.18\textsubscript{5.10} & 85.38\textsubscript{2.59} & 77.14\textsubscript{2.44} & 94.14\textsubscript{2.30} \\
1024 & \textbf{57.50\textsubscript{1.06}} & \textbf{87.31\textsubscript{1.59}} & \underline{79.78\textsubscript{2.73}} & \underline{94.89\textsubscript{1.12}} \\
\bottomrule
\end{tabular}
\caption{Accuracy of different stride values on four datasets.}
\label{tab:stride}
\end{table}

\section{Conclusion and Future Work}
In this paper, we propose DAG, a deformable attention-based graph neural network. By incorporating the actual spatial coordinates of patches, DAG constructs a dynamically weighted graph structure and introduces a deformable attention mechanism based on input features. Through the design of learnable offset modules, the model is able to dynamically capture structurally relevant regions from a global perspective, effectively modeling the complex tissue relationships inherent in pathological images. Extensive experiments and ablation studies conducted on four datasets demonstrate the effectiveness of DAG in recognizing pathological structures. 
In addition, we observe that DAG achieves strong performance in entity-centered diagnostic tasks, demonstrating a clear ability to delineate well-defined structural boundaries. In future work, we plan to focus on clinically relevant tasks that emphasize entity-level recognition and further extend our approach to downstream applications such as pathological image segmentation.

\bibliography{aaai2026}


\end{document}